\documentclass[final,12pt]{clear2023} 

\usepackage{caption}
\usepackage{wrapfig}

\newcommand\ind{\hspace{\algorithmicindent}}
\usepackage[normalem]{ulem}

\title[Enhanced Causal Discovery]{Enhancing Causal Discovery from Robot Sensor Data\\in Dynamic Scenarios}
\usepackage{times}
\usepackage{comment}
\usepackage{algorithm}
\usepackage{algorithmic}

\clearauthor{
    \Name{Luca Castri} \Email{lcastri@lincoln.ac.uk}\\
    \Name{Sariah Mghames} \Email{smghames@lincoln.ac.uk}\\
    \Name{Marc Hanheide} \Email{mhanheide@lincoln.ac.uk}\\
    \addr University of Lincoln, UK
    \AND
    \Name{Nicola Bellotto} \Email{nbellotto@dei.unipd.it}\\
    \addr University of Padua, Italy
}

\begin{document}

\maketitle

\begin{abstract}\label{sec:abstract}
Identifying the main features and learning the causal relationships of a dynamic system from time-series of sensor data are key problems in many real-world robot applications. In this paper, we propose an extension of a state-of-the-art causal discovery method, PCMCI, embedding an additional feature-selection module based on transfer entropy. Starting from a prefixed set of variables, the new algorithm reconstructs the causal model of the observed system by considering only its main features and neglecting those deemed unnecessary for understanding the evolution of the system. We first validate the method on a toy problem and on synthetic data of brain network, for which the ground-truth models are available, and then on a real-world robotics scenario using a large-scale time-series dataset of human trajectories. The experiments demonstrate that our solution outperforms the previous state-of-the-art technique in terms of accuracy and computational efficiency, allowing better and faster causal discovery of meaningful models from robot sensor data.
\end{abstract}

\section{Introduction} \label{sec:introduction}
Knowing the main variables contributing to the evolution of a dynamic system and their causal relationship is important for many real-world applications. For this reason, feature selection and causal discovery methods have become a crucial problem in machine learning and related areas, including robotics~\citep{scholkopf2021towards,Ahmed2021}.

Recently, the concept of causal analysis has extended over two fronts. \emph{Causal discovery} investigates the causal relationships between features of complex and dynamical systems to understand their evolution when their models are unknown. Indeed, in the past few decades, many causal discovery algorithms for static and time-series data have been developed~\citep{glymour_review_2019}.
Another branch of causality/machine learning, \emph{feature selection}, tries to identify the most meaningful set of variables that is responsible for the evolution of the system~\citep{chandrashekar2014survey}. \emph{Causal representation} can be considered the meeting point between these two fronts. It extrapolates the relevant features characterising the observed system and reconstructs a causal model between them \citep{yao2022temporally,lippe2022citris,lachapelle2022disentanglement,wang2022causal}.

\begin{wrapfigure}{r}{0.6\textwidth}
\centering
\includegraphics[trim={0cm 0cm 4.5cm 4cm}, clip, width=0.6\textwidth]{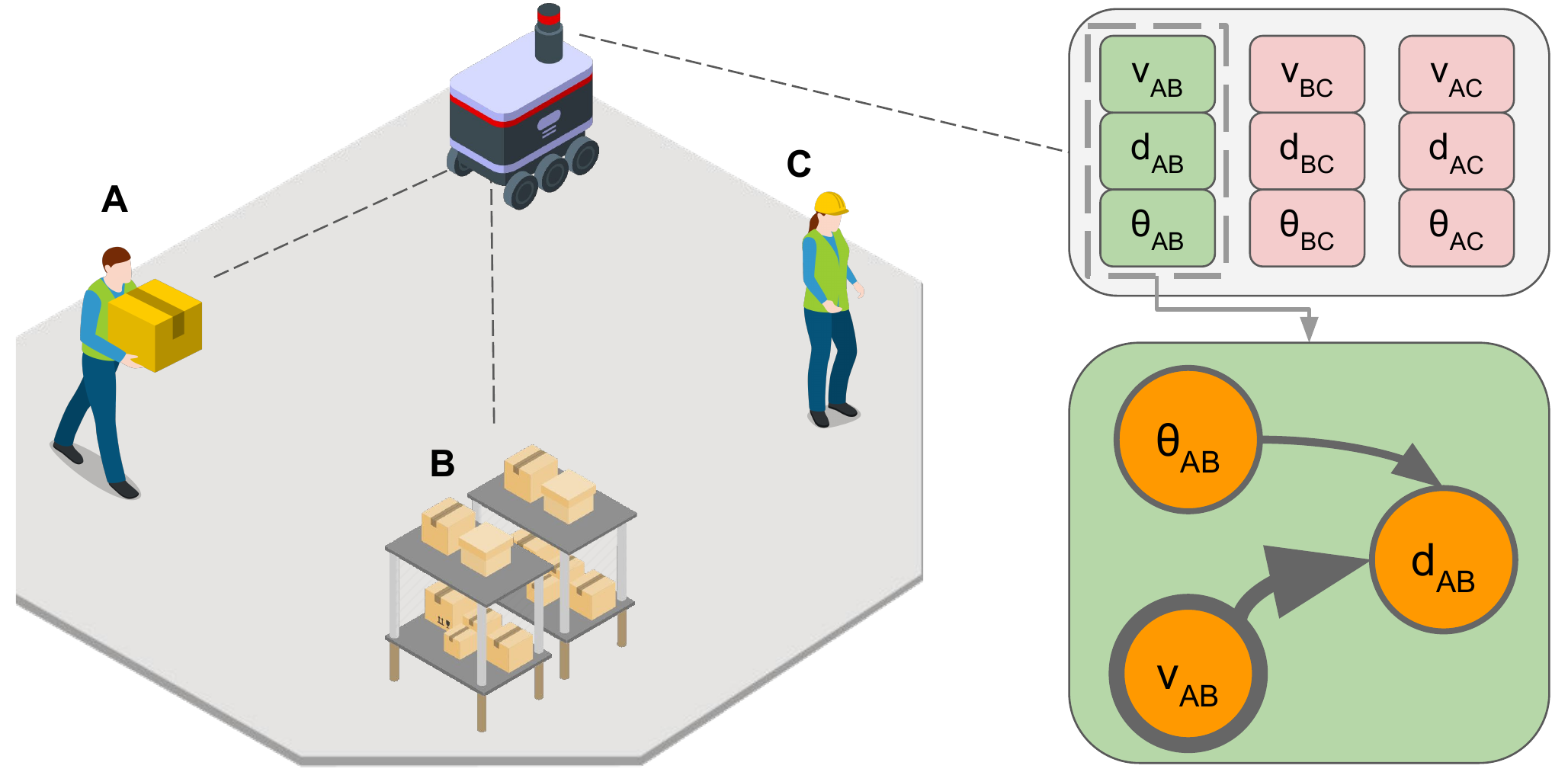}
\caption{
A mobile robot in a warehouse-like environment observes the interaction between agents A and B. By using our approach, the robot can disregard the interactions AC and BC as agent B is a static object and agent C is a standing human, not involved in the interaction.}
\vspace{-10pt}
\label{fig:intro}
\end{wrapfigure}

To our knowledge though, none of these approaches extracts both the important features representing the system and the causal association between them, while at the same time taking into account the execution time and the computational cost for completing the task. Causal analysis of complex and dynamical systems is extremely demanding in terms of time and hardware resources \citep{runge2020discovering, wienobst2021extendability}, making it a challenge for autonomous robotics with limited hardware resources and real-time requirements \citep{castri2023continual}.
We therefore aim to extend one of the state-of-the-art causal discovery methods, i.e. PCMCI~\citep{runge_causal_2018}, augmenting it with a feature-selection algorithm that is able to identify the correct subset of variables to involve in the causal analysis, starting from a prefixed set of them. Hence, we introduce an all-in-one approach that identifies the causal features representing the system and, based on them, builds a causal model directly from time-series data. As a consequence, the causal discovery process turns out faster and more accurate.
For instance, in an automated warehouse scenario (see Fig.~\ref{fig:intro}), where a robot observes the interactions among objects and humans (e.g. worker and shelf),
it is important to know which features, among those detectable by the robot's on-board sensors, are relevant for describing the observed interaction (e.g. human-shelf distance/angle, human velocity, etc.), and which instead can be neglected (e.g. other humans not involved in the interaction). Our approach would allow the robot to discard unnecessary features and build a causal model of the interaction using only those actually involved in the process.

In this preliminary work, we demonstrate that by starting from a prefixed set of observable variables and applying a suitable feature selection method based on transfer entropy, our algorithm is able to extract those that better characterise the evolution of the system and find the causal relationships among them, significantly faster and more accurately than PCMCI.
In summary, our contributions are the following:
\begin{itemize}
  \setlength{\itemsep}{1pt}
  \setlength{\parskip}{0pt}
  \setlength{\parsep}{0pt}
    \item an all-in-one algorithm able to select the most meaningful features from a set of variables and build a causal model from such selection. To this end, we significantly enhance speed and accuracy of the causal discovery;
    \item experimental evaluation of the algorithm and the obtained causal models on a challenging robotics dataset to predict spatial interactions in dynamic environments.
\end{itemize}

The paper is structured as follows: related work about feature selection, causal discovery and causal representation learning are presented in Sec.~\ref{sec:related_work}; Sec.~\ref{sec:TE} introduces the motivation for using a transfer entropy-based strategy in our approach; Sec.~\ref{sec:approach} explains the implementation details of our method; Sec.~\ref{sec:experiments} presents simulation and real-world results to validate, respectively, correctness and performance of our solution; finally, we conclude the paper in Sec.~\ref{sec:conclusion} discussing achievements and future work.

\section{Related Work} \label{sec:related_work}
\textbf{Feature selection:}
The increasing use of machine learning for high-dimensional data analysis led to the development of several feature selection methods. Indeed, feature selection helps to overcome the high dimensionality challenge by eliminating redundant and irrelevant data. Removing the irrelevant features improves learning accuracy and reduces the computation time. Feature selection methods can be categorized in three different categories, reviewed in~\citep{chandrashekar2014survey}. The first, {\em filter methods}, use general statistics metrics (e.g. correlation, mutual information) to select the most useful features; the second, {\em wrapper methods}, use a predictor as a black box and its performance as the objective function to evaluate the variable subset; finally, the third, {\em embedded methods}, integrate the feature selection process into a classification/regression model.
For their capability to deal with high-dimensional data and low computational cost, filter methods are often used as feature selection ones. In particular, the Transfer Entropy~(TE)~\citep{PhysRevLett.85.461,Bossomaier2016AnIT} is an extension of mutual information that measures the directed information transfer between time series of a source and a target variable. TE has become popular in many scientific disciplines to infer dependencies and whole networks from data. Recently, it has often been used as a causal discovery method~\citep{zeng2022satellite,chen2021satellite,zeng2022spacecraft}, although~\cite{runge2012quantifying} and \cite{janzing2013quantifying} demonstrated that TE does not satisfy the full set of principles that a causal measure must satisfy, as discussed later in Sec.~\ref{sec:TE}.
However, for its capability to find the parents of each target variable, TE has been adopted in this paper as feature selection method to have an initial causal model preview.\\[0.1cm]
\noindent \textbf{Causal discovery:}
Structural causal models (SCMs) are at the core of causal discovery~\citep{alma991011292629705181}, complemented by opportune Directed Acyclic Graphs~(DAGs) with nodes and oriented edges to represent, respectively, system variables and dependencies between them.
The knowledge of SCMs leads to the possibility to reason on the cause and effect relationship between variables. 
Several methods have been developed over the last few decades to derive causal relationships from observational data. They can be categorized into two main classes, reviewed in~\citep{glymour_review_2019}. The first, {\em constraint-based methods}, such as Peter and Clark~(PC) and Fast Causal Inference~(FCI), rely on conditional independence tests as constraint-satisfaction to recover the causal graph, while the second, {\em score-based methods}, such as Greedy Equivalence Search (GES), assign a score to each DAG and perform a search in this score space. More recently, reinforcement learning-based methods have also been used to discover causal structure~\citep{Zhu2019}.
However, most of these algorithms work only with static data~(i.e. no temporal information), which is a limitation in many robotics applications.
Indeed, methods for time-dependent causal discovery are necessary to deal with time-series of sensor data. To this end, a variation of the PC algorithm, called PCMCI, was adapted and applied to time-series data~\citep{runge_causal_2018,runge_detecting_2019,saetia_constructing_2021}.


%
In robotics, \cite{brawer_causal_2021} presents a method to build and learn a SCM through a mix of observation and self-supervised trials for tool affordance with a humanoid robot.
Other applications include the use of PCMCI to derive the causal model of an underwater robot trying to reach a target position~\citep{cao_reasoning_2021} or to predict human spatial interactions in a social robotics context~\citep{castri2022causal}.
Further causality-based approaches can be found in the robot imitation learning and manipulation area~\citep{Katz2018,Angelov2019,Lee2022}.
However, all these solutions rely on a pre-determined set of variables for performing the causal analysis and do not extract the most meaningful ones for the reconstruction of the causal model. In real-world scenarios, including the robotics sector, such set of variables might be extremely large, leading to slow causal analysis due to the computational cost, which increases with the number of variables, and to inaccurate causal models with a high percentage of spurious links.\\[0.1cm]
\textbf{Causal representation:}
In many real-world scenarios, not all the features characterising a system are observable and those observed can be affected by latent temporal processes and confouders. Recent causal representation approaches deal with latent variables. 
\cite{yao2022temporally} aims to recover time-delayed latent causal variables and identify their relations from measured temporal data under stationary and non-stationary environments. The recovered quantities need then to be interpreted by a human expert as high-level physical variables. 
\cite{lippe2022citris} and \cite{lachapelle2022disentanglement} exploit pre- and post-intervention observations in adjacent time frames to study causal relationships between latent variables.
The strategy of \citep{wang2022causal} is relatively close to our approach: auto-dependent variables with no link from/to other variables are kept and will appear isolated in the final causal model. In addition to this though, our approach can also remove the variables that are not auto-dependent and do not have any other link from/to other variables (e.g. noise).

All these works, however, aim to increase the causal model accuracy with no consideration for the execution time. Instead, our approach does not deal directly with latent variables but aims to build, in a reasonable amount of time, an accurate causal model starting from a prefixed set of observed variables and removing the unnecessary ones to simplify the causal discovery computation. Another key difference between all these approaches and ours is that the latter does not need a training procedure, since it is based on conditional independence tests between variables. Moreover, most of those approaches need active interventions, while ours is based on observational data, although the quality of the causal analysis could be improved by providing also interventional data.

\section{Transfer Entropy-based Filter}\label{sec:TE}
As already explained in the Sec.~\ref{sec:related_work}, the Transfer Entropy~(TE) ~\citep{PhysRevLett.85.461,Bossomaier2016AnIT} is an extension of mutual information that measures the directed information transfer between time-series of a source and a target variable. 
It represents the information-theoretic analog of Granger causality~\citep{barnett2009granger}. TE has become popular in many scientific disciplines 
as a causal discovery method~\citep{chen2021satellite,zeng2022satellite,zeng2022spacecraft}. 

Nevertheless, \cite{runge2012quantifying} and \cite{janzing2013quantifying} demonstrated the limitations and the problems of TE as a causal measure. In particular, \cite{runge2012quantifying} established five key criteria that such a measure should meet:
\begin{itemize}
  \setlength{\itemsep}{1pt}
  \setlength{\parskip}{0pt}
  \setlength{\parsep}{0pt}
    \item \textit{generality}: the measure should not be restricted to certain types of associations (e.g. linear);
    \item \textit{equitability}: the measure should reflect
    a certain heuristic notion of coupling strength, i.e., it should give similar scores to equally noisy dependencies;
    \item \textit{causality}: the measure should give nonzero values only to dependencies among components of a multivariate process that are not conditionally independent given the remaining process;
    \item \textit{coupling-strength autonomy}: the measure is uniquely determined by the interaction between the two components under examination, independently from their interaction with the remaining part of the process;
    \item \textit{practically computable}.
\end{itemize}
Moreover, \cite{runge2012quantifying} introduced a causal measure related to TE, called \textit{momentary information transfer}~(MIT), which represents the base for the PCMCI causal discovery algorithm~\citep{runge_causal_2018}. They showed that MIT satisfies all the above-mentioned principles, unlike TE.

To clarify the difference between the two measures, let's consider a process $Z$ composed by three sub-processes: $X$, $Y$ and $W$. TE and MIT from $X$ to $Y$ can be defined as differences of conditional entropies:
\begin{equation}
    I^{TE}_{X\rightarrow Y} = H(Y_t \mid Z_t^-\smallsetminus X_t^-) - H(Y_t \mid Z_t^-)
    \label{eq:TE}
\end{equation}
\begin{equation}
    I^{MIT}_{X\rightarrow Y}(\tau) = H(Y_t \mid P_{Y_t} \smallsetminus \{X_{t-\tau}\}, P_{X_{t-\tau}}) - H(Y_t \mid P_{Y_t})
    \label{eq:MIT}
\end{equation}
where (\ref{eq:TE}) quantifies $Y$'s entropy reduction, computed at time $t$, when the infinite past of $X_t^-$ is included in the conditioning set $Z_t^-$ (i.e. the infinite past of the whole process). 
The measure of (\ref{eq:MIT}) is related to (\ref{eq:TE}) but, in this case, \emph{only} the parents of the variables involved in the test at a specific time lag are considered and not the the infinite past of the complete conditioning set.
%

Both (\ref{eq:TE}) and (\ref{eq:MIT}) show that TE is not a proper measure of causal strength, therefore it is not suitable for causal discovery.
There are three main limitations.
I)~Unlike MIT, which computes the influence of $X$ to $Y$ at a specific time $t-\tau$, TE is not lag-specific. This can lead to false interpretations, like in the case the system under examination has feedback.
II)~Since TE is not lag-specific, but considers all the past lags of the process, it is not computable in practice.
A workaround for that is the use of a truncation parameter to discard the infinite past lags of the process and maintain only the lags up to the maximal coupling delay of the variable involved in the measure. However, the introduction of this workaround has a strong influence on the TE value and affects its reliability.
III)~Finally, in the computation of the causal strength from $X$ to $Y$, TE is not uniquely determined by the interaction of the two components alone but considers the whole process $Z$ (thus also including $W$). Therefore, it is influenced by the misleading effects of, e.g., auto-dependency and interaction with other processes, while MIT considers only the parents of the processes involved.

Essentially, TE does not fulfill the criteria of \emph{coupling-strength autonomy} and \emph{practically computable}, so it can not be a measure of causal strength. On the other hand though, MIT requires knowing the parents of the variables involved. This makes MIT 

\begin{wraptable}{r}{0.35\textwidth}
\vspace{-11pt}
\centering
\begin{tabular}{c|c|c}
\multicolumn{1}{l|}{}      & \multicolumn{1}{l|}{MIT}  & \multicolumn{1}{l}{TE}    \\ \hline
generality                 & \checkmark & \checkmark \\
equitability               & \checkmark & \checkmark \\
causality                  & \checkmark & \checkmark \\
coupling-strength & \checkmark &                           \\
computable     & \checkmark & $\circ$
\end{tabular}
\caption{MIT/TE comparison: ``\checkmark'' indicates a criterion is met; ``$\circ$'' indicates a criterion can be satisfied with a workaround.}
\label{table:TEvsMIT}
\end{wraptable}
\noindent not suitable for feature selection. Indeed, the PCMCI algorithm first performs a PC step to determine an initial causal structure, and then it exploits the latter during a MCI step (based on MIT). 
Table~\ref{table:TEvsMIT} recaps the MIT/TE comparison.

Despite the above limitations, TE is still a valid choice for the feature selection problem. Due to the unmet coupling-strength autonomy criteria, it is not an accurate causal measure, but it can still indicate whether a relation between two variables exists or not (causality principle). For this reason, similarly to~\citep{wollstadt2018idtxl}, we adopted TE as feature selection method for our approach.

\section{Filtered-based Causal Discovery} \label{sec:approach}
\begin{figure}[!t]
\noindent\begin{minipage}[t]{.54\textwidth}
\begin{algorithm}[H]
\small
\caption{F-PCMCI} \label{alg:fpcmci_pseudocode}
\begin{algorithmic}[1]
\REQUIRE{
    time-series data $D$, significance threshold $\alpha$,\\
    \hspace{0.8cm} min and max time lag $\tau_{min}$, $\tau_{max}$
}

    \STATE $CS = \{\}$ \hspace{\fill}{\footnotesize$\leftarrow$ hypothetical causal structure dictionary}
    \STATE \textbf{for each} target $T$ in $D$ \textbf{do}
        \STATE\ind $S_T=\emptyset$ \hspace{\fill}{\footnotesize$\leftarrow$ T sources / conditioning set}
        \STATE\ind $L = [~]$ \hspace{\fill}{\footnotesize$\leftarrow$ temporary list}
        \STATE\ind \textbf{while} $D$ not empty \textbf{do}
            \STATE\ind\ind \textbf{for each} source $S$ in $D \smallsetminus T$ \textbf{do}
                \STATE\ind\ind\ind $(p$-$value, I)_{S}$ = TE$_{S \rightarrow T|S_T}(\tau_{min}, \tau_{max})$
                \STATE\ind\ind\ind add $(p$-$value, I)_{S}$ to $L$ 
            \STATE\ind\ind $(p$-$value, I)_{S_b} = \arg \max_{I}(L)$ \hspace{\fill}{\footnotesize$\leftarrow$ best candidate}
            \STATE\ind\ind \textbf{if} $p$-$value \leq \alpha$ \textbf{then}
                \STATE\ind\ind\ind remove $S$ from $D$ and add $S$ to $S_T$
            \STATE\ind\ind \textbf{else}
                \STATE\ind\ind\ind \textbf{if} $S_T \neq \emptyset$ \textbf{then} $CS(T) = S_T$
                \STATE\ind\ind\ind \textbf{break}
\STATE $D_s \leftarrow$ shrink original D by $var_{sel}=$\texttt{keys}(CS)
\STATE CM = \texttt{PCMCI}($D_s$, $\alpha$, $\tau_{min}$,  $\tau_{max}$, CS)
\RETURN CM \hspace{\fill}{\footnotesize$\leftarrow$ causal model}
\end{algorithmic}
\end{algorithm}
\end{minipage}%
\hspace{0.15cm}
\begin{minipage}[t]{.45\textwidth}
  \centering
    \strut\vspace*{-\baselineskip}\newline\includegraphics[trim={7.5cm 0.75cm 7.5cm 0cm}, clip, width=0.92\textwidth]{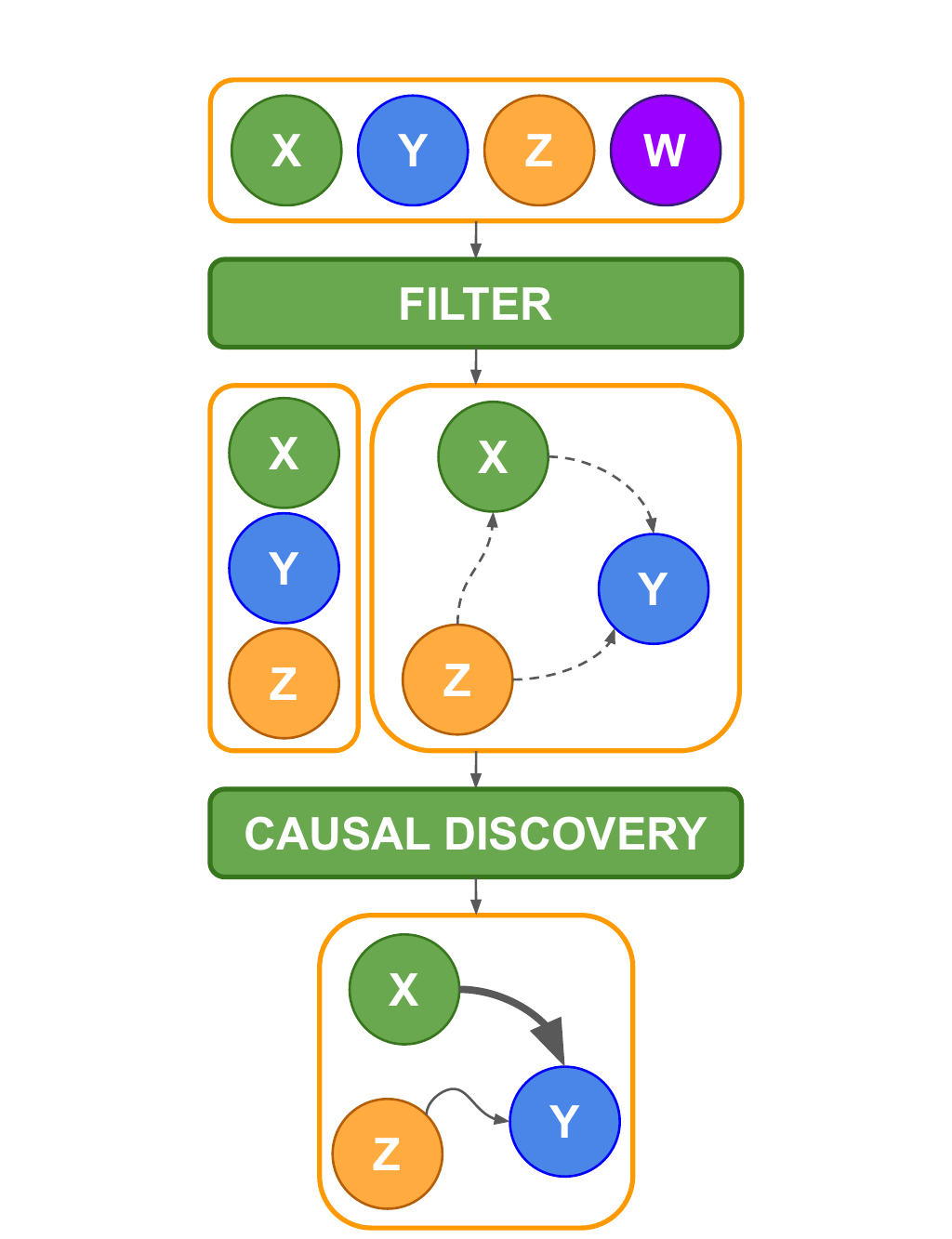}
    \captionof{figure}{F-PCMCI block-scheme representation with an example.}
    \label{fig:fpcmci_flowchart}
\end{minipage}
\end{figure}
\noindent Our approach, named Filtered PCMCI~(F-PCMCI), uses a TE-based method to "filter" the important features and their possible associations from the whole set of variables, before the actual causal analysis. A Python implementation of F-PCMCI has been developed and made publicly available\footnote{\url{https://github.com/lcastri/fpcmci}}. As explained in Sec.~\ref{sec:TE}, we used TE to decide which variables and links can be excluded from the original set, and those which are needed for the causal analysis. As output, the filter returns a set of variables and a hypothetical causal model, which then needs to be validated by a proper causal analysis. The latter is performed by the PCMCI causal discovery algorithm, briefly explained in the following paragraph. A pseudo-code implementation and a block diagram of our approach are also illustrated in Alg.~\ref{alg:fpcmci_pseudocode} and Fig.~\ref{fig:fpcmci_flowchart}, respectively. The latter shows an example including the main F-PCMCI operations: from a set of variables ($X,~Y,~Z,~W$), the Filter block removes unnecessary variables ($W$) and feeds into the causal discovery block only the remaining ones ($X,~Y,~Z$) with a hypothetical causal structure.

Note that the ability of F-PCMCI to remove unnecessary variables does not create any kind of accuracy problem in the reconstruction of the causal model, even if there are hidden confounders. In the latter case, since both F-PCMCI and PCMCI need a pre-determined set of variables to start the causal analysis, if a confounder is not in the initial set, it will not be considered. Therefore, the two algorithms would behave exactly the same. For example, if the confounder $Z$ of two variables $X$ and $Y$ is not included in the initial set, a spurious correlation will be detected between $X$ and $Y$. Thus, since F-PCMCI removes only variables that have no links to other variables (including links to themselves), they cannot be removed.
\\[0.1cm]
\noindent\textbf{PCMCI:} this is a causal discovery algorithm~\citep{runge_causal_2018} that consists of two main parts, both exploiting conditional independence tests (e.g. partial correlation, Gaussian processes, and distance correlation) to measure the causal strength between variables. The first part is the well-known PC algorithm~\citep{glymour_review_2019}, which performs the following procedure to reconstruct an initial version of the causal model structure: \textit{(i)} start from a fully connected and undirected graph; \textit{(ii)} remove edges between variables that are unconditionally independent; \textit{(iii)} for each pair of variables ($A, B$) with an edge between them, and for each variable $C$ with an edge connected to either of them, remove the edge between $A$ and $B$ if $A \bot B \mid C$; \textit{(iv)} for each pair of variables ($A, B$) with an edge between them, and for each pair of variables ($C, D$) with edges connected to both $A$ or $B$, eliminate the edge between $A$ and $B$ if $A \bot B \mid (C, D)$; finally, \textit{(v)} apply the \emph{v-structure} and the \emph{orientation propagation} rules to orient all the edges. At this stage, the second part of PCMCI, i.e. the MCI test, validates the estimated structure by computing the test statistics and p-values for all the links, and then outputs the final causal model.

\subsection{Toy Problems} \label{subsec:toyproblem_appr}
To evaluate the correctness of the approach, two toy problems with known ground-truth causal models were considered. The first one~(\ref{eq:tau1}), hereinafter called~$S_1$, is a nonlinear system of equations with a maximum time lag of 1. The second one (\ref{eq:tau2}), called~$S_2$, is a nonlinear system with a maximum time lag of 2.
In both cases, the structure of the nonlinear systems were defined in advanced as follows:
\noindent\begin{minipage}[t]{.49\textwidth}	 
\begin{equation}
S_1 = 
    \begin{cases} \label{eq:tau1}
      x_{0_t} = c_{00}x_{0_{t-1}} - c_{01}x_{1_{t-1}}c_{02}x_{2_{t-1}}+ \eta_{0_t}\\
      x_{1_t} = \eta_{1_t}\\
      x_{2_t} = \frac{c_{21}x_{1_{t-1}}}{1+c_{22}x_{2_{t-1}}} + \eta_{2_t}\\
      x_{3_t} = c_{33} + \sqrt{x_3(t-1)} + \eta_{3}(t)\\
      x_{4_t} = \frac{c_{41}x_{1_{t-1}}c_{42}x_{2_{t-1}}}{1 + c_{43} x_{3_{t-1}}} + \eta_{4_t}\\
      x_{5_t} = \eta_{5_t}\\
      x_{6_t} = \frac{c_{60}x_{0_{t-1}}}{1 + c_{65}x_{5_{t-1}}} + \eta_{6_t}
    \end{cases}
\end{equation}
\end{minipage}
\begin{minipage}[t]{.49\textwidth}	 
\begin{equation}
S_2 = 
    \begin{cases} \label{eq:tau2}
      x_{0_t} = c_{01}x_{1_{t-2}}c_{02}x_{2_{t-1}} + \eta_{0_t}\\
      x_{1_t} = \eta_{1_t}\\
      x_{2_t} = c_{21}x_{1_{t-2}}^2 + \eta_{2_t}\\
      x_{3_t} = c_{33} + x_{3_{t-1}} + \eta_{3_t}\\
      x_{4_t} = c_{42}x_{2_{t-2}} - c_{43}*x_{3_{t-1}} + \eta_{4_t}\\
      x_{5_t} = \frac{c_{50}x_{0_{t-1}}}{1 + c_{55}x_{5_{t-2}}} + \eta_{5_t}\\
      x_{6_t} = \eta_{6_t}
    \end{cases}
\end{equation}
\end{minipage}\\[0.2cm]
where $x$ represents the system variables and $c$ the coefficients. These equations were chosen to test various types of dependencies, including linear and non-linear cross- and auto-dependencies, including noise with and without other relationships, and different time-lag dependencies.
The chosen noise $\eta$ has a uniform distribution with range~$[0, 1)$ for both toy problems. Finally, to test the ability of F-PCMCI to detect different ranges of link strength, the coefficient $c$ was also assigned a uniform distribution with range $[0,1)$ for $S_1$ and $[0,10)$ for $S_2$.

\subsection{Modeling Real-world Human Spatial Interactions} \label{subsec:realworld_appr}
We used our approach to model and predict spatial interactions (Fig.~\ref{fig:exp_setup} left). This application involves three main steps: (\textit{i})~extracting time-series of sensor data from human spatial interaction scenarios; (\textit{ii})~reconstruct the causal model using F-PCMCI; (\textit{iii})~embedding the causal model in a LSTM-based prediction system. Our implementation of the latter\footnote{\url{https://github.com/lcastri/cmm\_ts}} is inspired by~\citep{yin2021multi}, which uses an encoder-decoder network with two attention mechanisms in the encoder block and one in the decoder network. In particular, the encoder block includes an input attention mechanism that selects the most useful time steps of the considered observation window 
, plus a self-attention mechanism that selects the most meaningful drivers for predicting the target variable. 
Since the focus of this approach is to highlight the benefit of having a causal-informed input to the network, we removed the decoder temporal attention layer of the original architecture.

Moreover, for its ability to select the drivers for each target variables, the self-attention mechanism was used to integrate our discovered causal model as follows:
\begin{center}
\begin{minipage}{.49\textwidth}
\begin{equation}
    g_t = \tanh(W_gx_t + b_g)
\end{equation}
\end{minipage}
\begin{minipage}{.49\textwidth}
\begin{equation}
    \alpha^k_t = \sigma(W_\alpha g_t + b_\alpha)
\end{equation}
\end{minipage}
\end{center}
where $\sigma$ denotes the sigmoid function, $W_g$ and $W_\alpha$ are the learnable parameters, $b_g$ and $b_\alpha$ are bias vectors, and $x_t$ represents the input driving series at time step $t$ for the target variable $x^k$. We used the bias vector $b_\alpha$ to embed the causal inference vector extracted from our causal model corresponding to the target variable $x^k$, and we set it as a non-trainable parameter for the network.



\section{Experiments} \label{sec:experiments}
To evaluate our approach and verify its advantages in terms of computational cost and causal models' accuracy with respect to PCMCI, we first validate it with the toy problems described in Sec.~\ref{subsec:toyproblem_appr} and with Functional Magnetic Resonance Imaging~(fMRI) time-series data generated with a tool\footnote{\url{https://www.fmrib.ox.ac.uk/datasets/netsim/}} provided by~\citep{smith2011network}. The latter is able to generate realistic and rich simulations of fMRI time-series data with ground-truth brain networks. Once, established that our approach works correctly, we used it for modeling and predicting human spatial behaviours (using the network described in Sec.~\ref{subsec:realworld_appr}) on a challenging dataset, i.e. TH{\"O}R~\citep{thorDataset2019}. The latter contains data of people moving in an indoor environment, arranged as a workshop/warehouse. Our strategy is first to extract the real sensor time-series data from the dataset, as explained in Sec.~\ref{sec:approach}, and then use them for causal discovery. The effectiveness of our approach is demonstrated by comparing causal and non-causal predictions. A further comparison between PCMCI and F-PCMCI is provided to illustrate the advantages of our method with respect to the state-of-the-art.

\subsection{Evaluation on Toy Problems}\label{subsec:toy_results}
\begin{figure}[!h]
\centering
    \includegraphics[trim={0.5cm 0.5cm 1.3cm 1.1cm}, clip, scale=0.55]{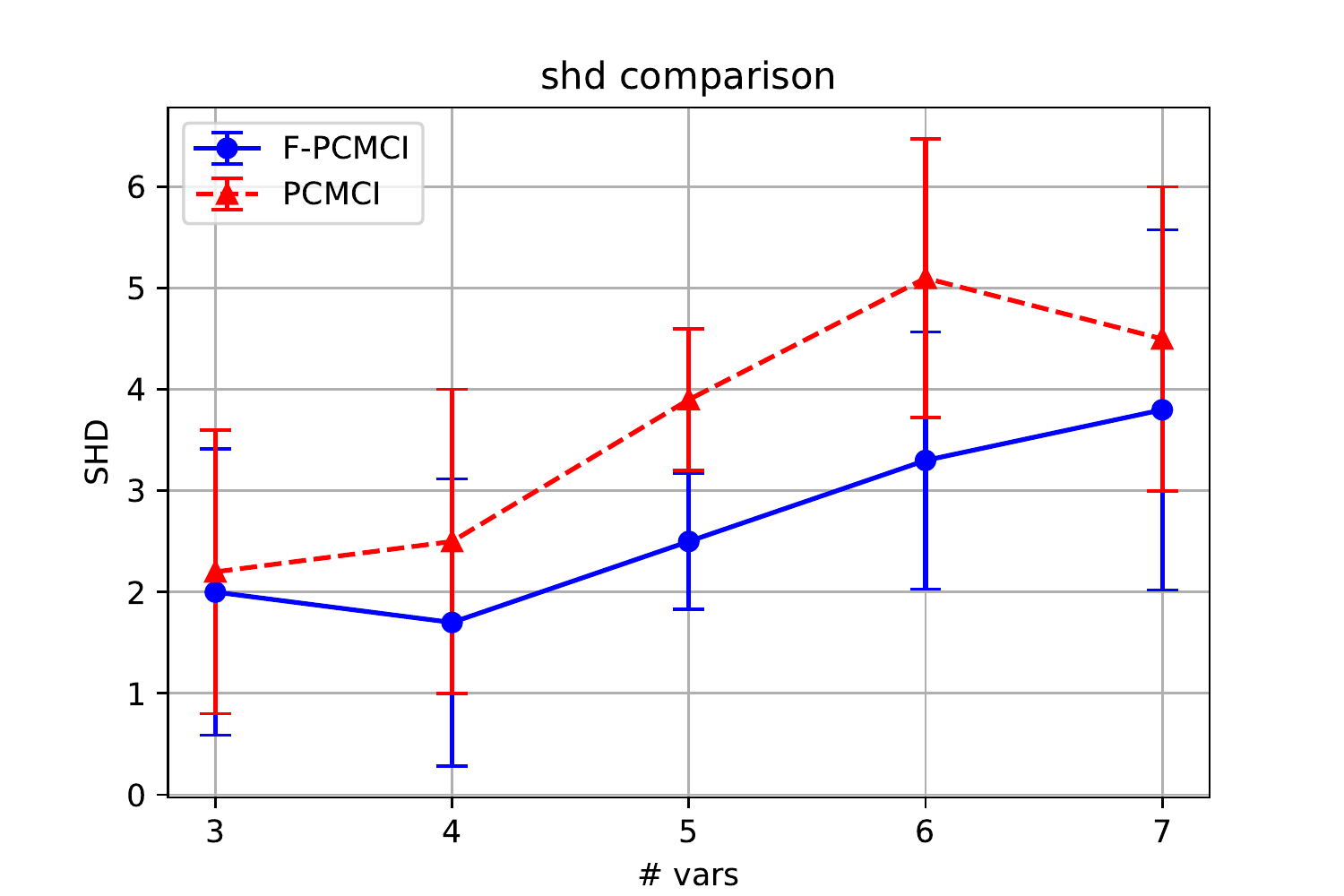}
    \label{fig:exp_S1_shd}
    \includegraphics[trim={0.5cm 0.5cm 1.3cm 1.1cm}, clip, scale=0.55]{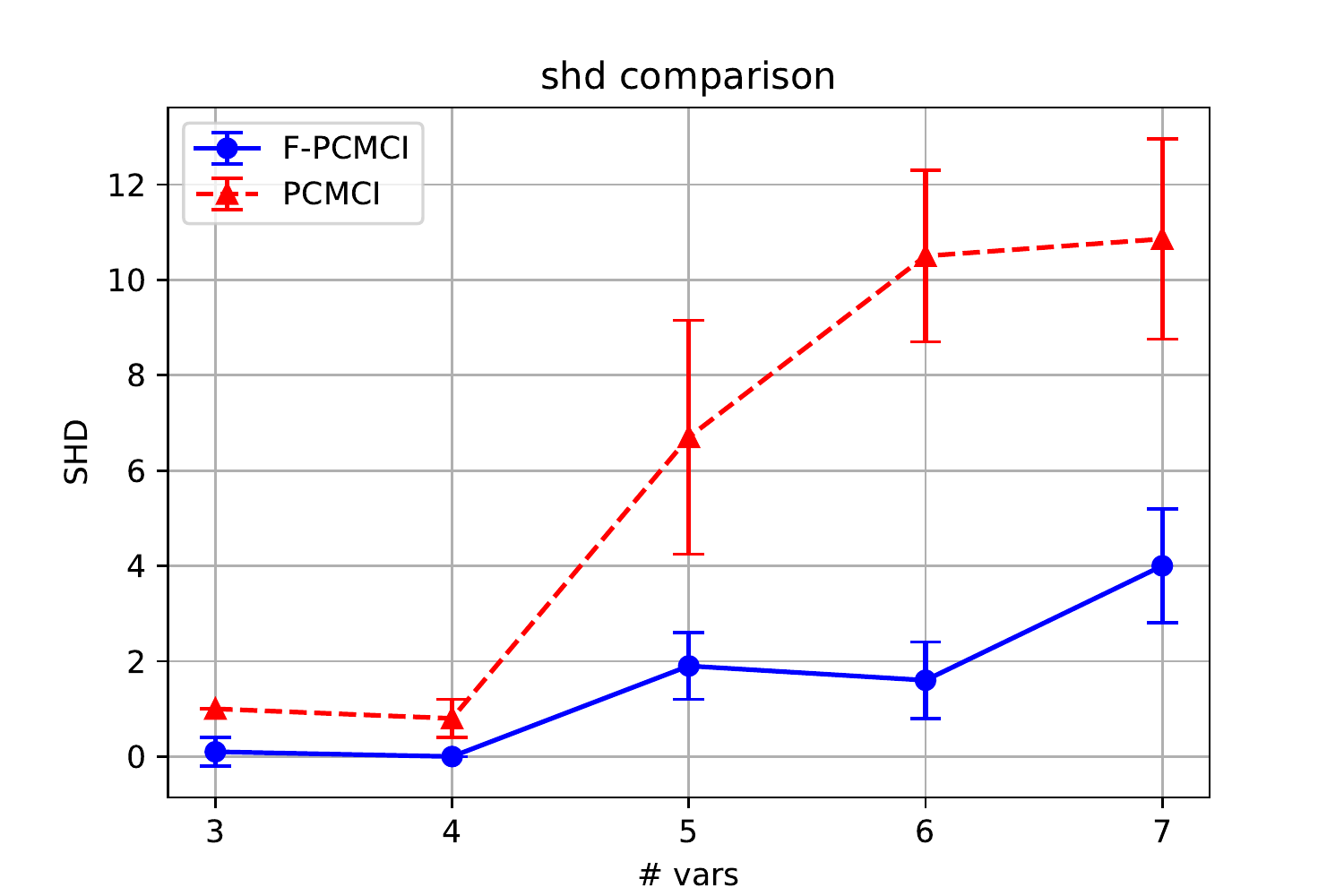}
    \label{fig:exp_S2_shd}
    \\
    \includegraphics[trim={0.5cm 0.5cm 1.3cm 1.1cm}, clip, scale=0.55]{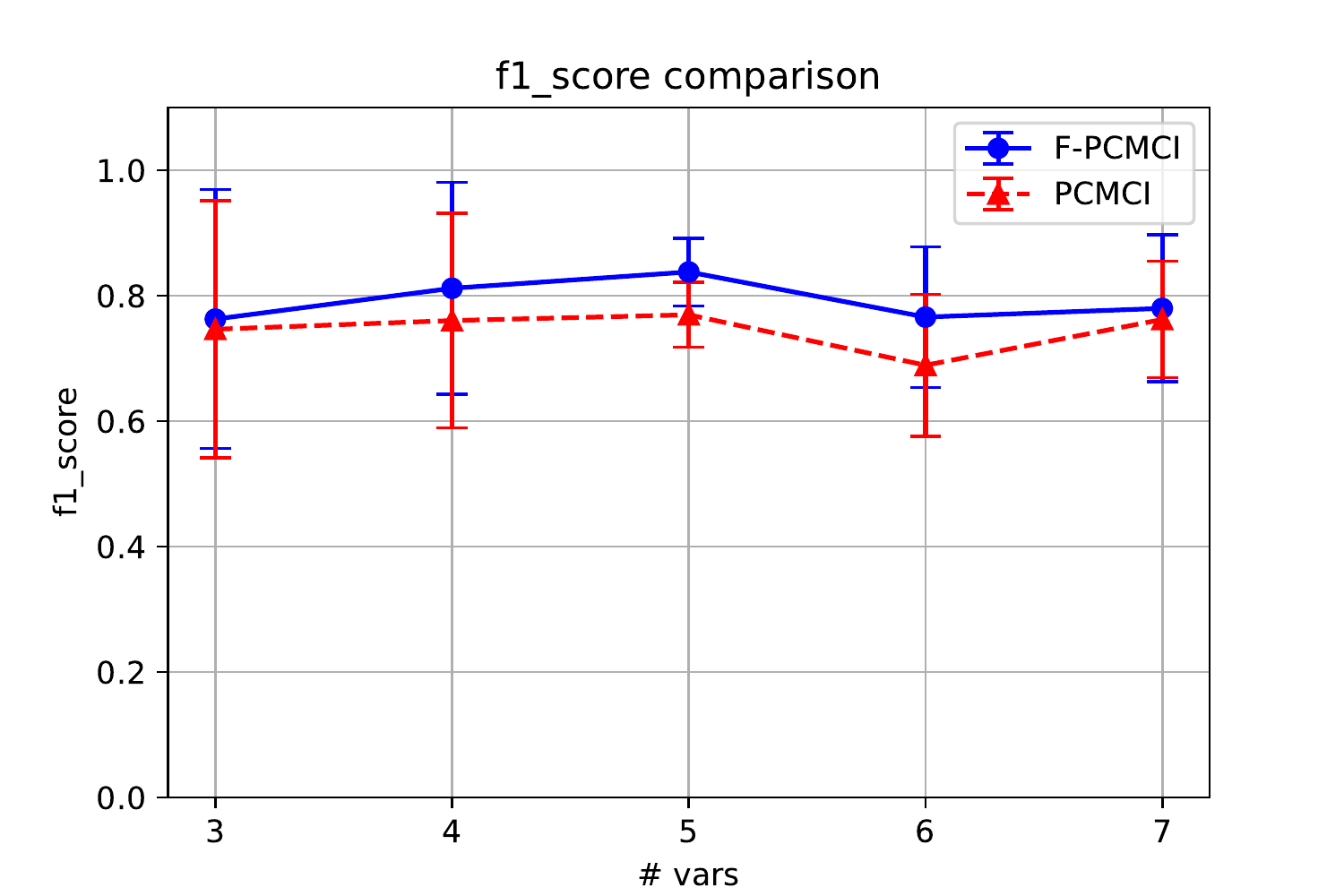}
    \label{fig:exp_S1_f1_score}
    \includegraphics[trim={0.5cm 0.5cm 1.3cm 1.1cm}, clip, scale=0.55]{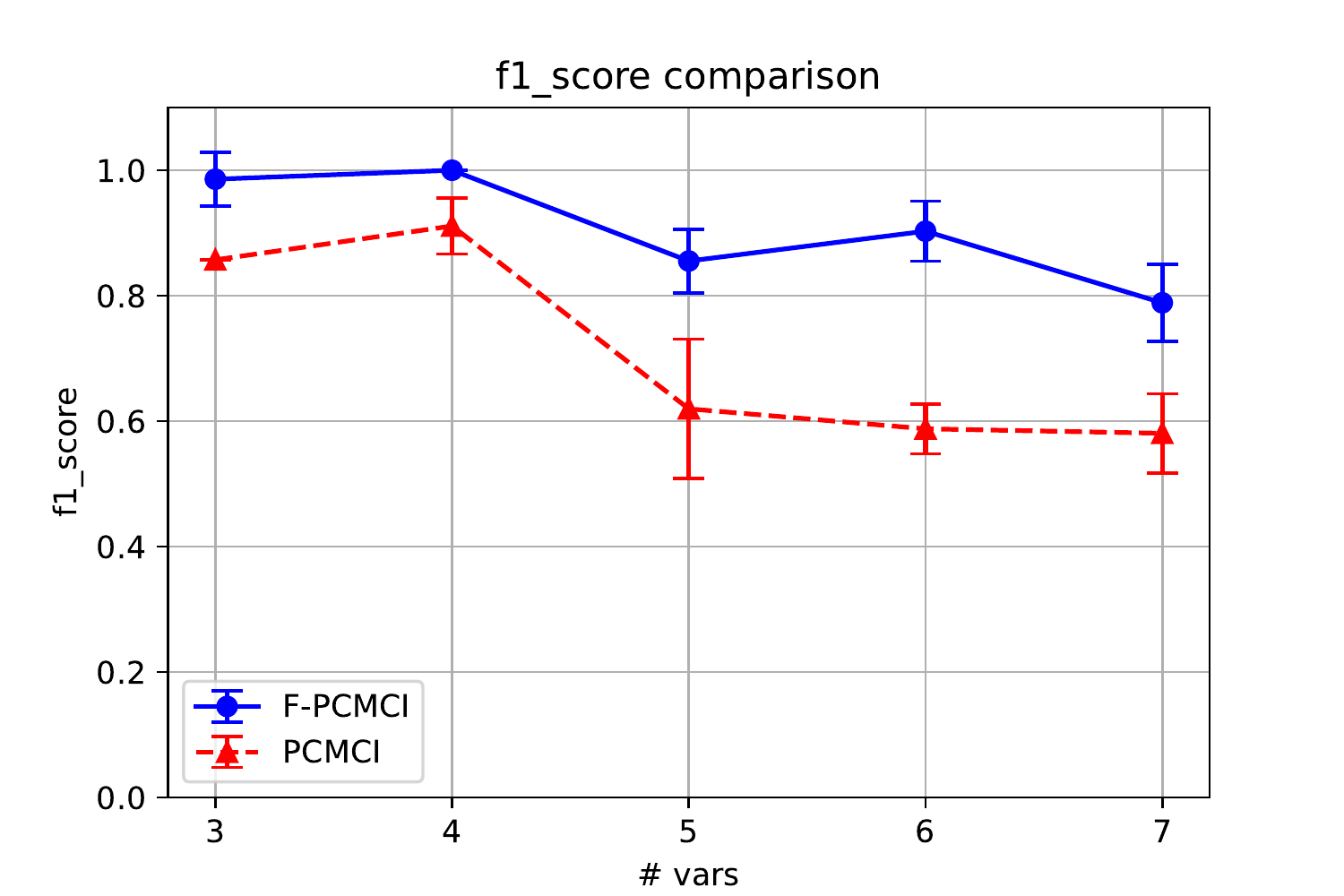}
    \label{fig:exp_S2_f1_score}\\
    \includegraphics[trim={0.5cm 0.1cm 1.3cm 1.1cm}, clip, scale=0.55]{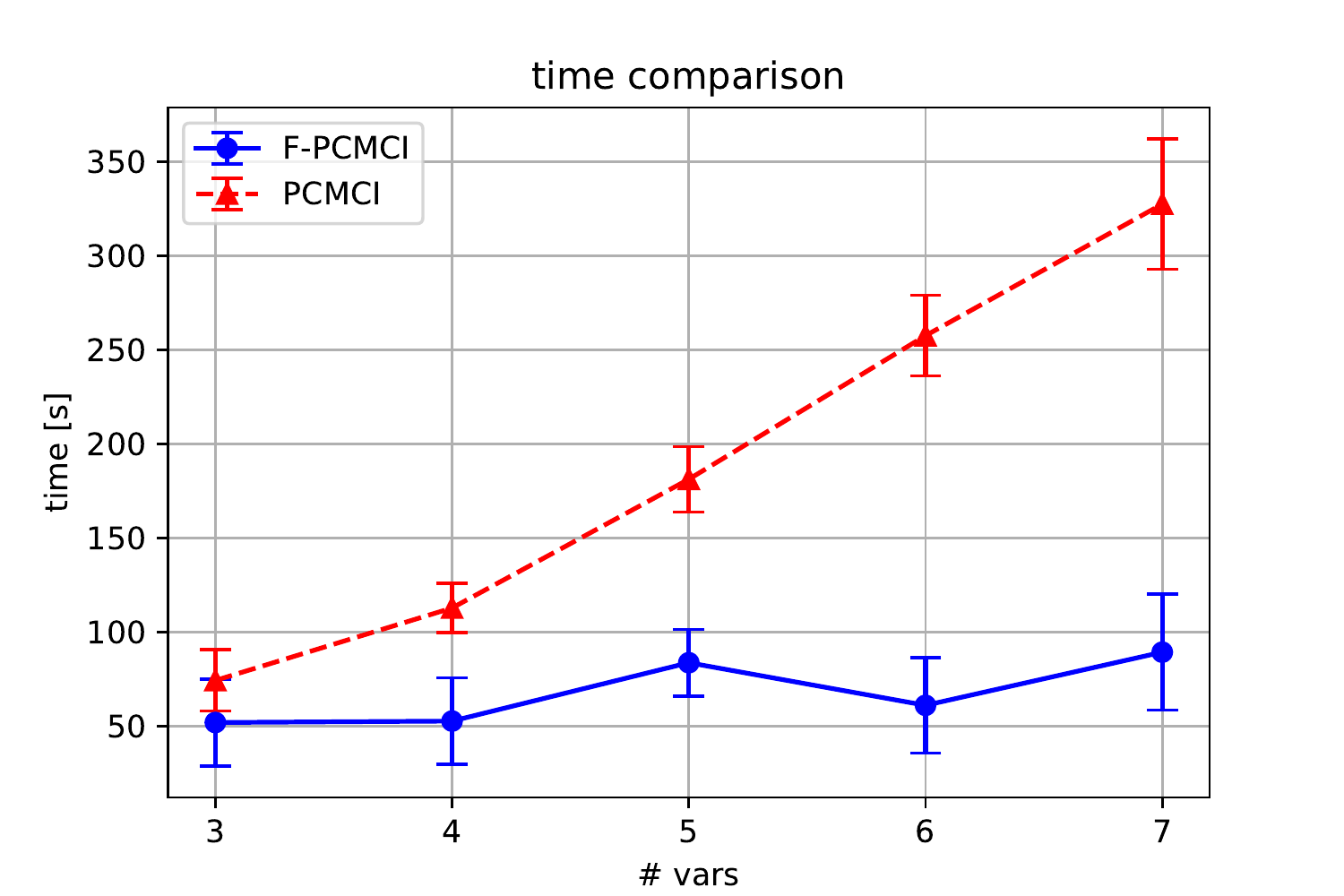}
    \label{fig:exp_S1_time}
    \includegraphics[trim={0.5cm 0.1cm 1.3cm 1.1cm}, clip, scale=0.55]{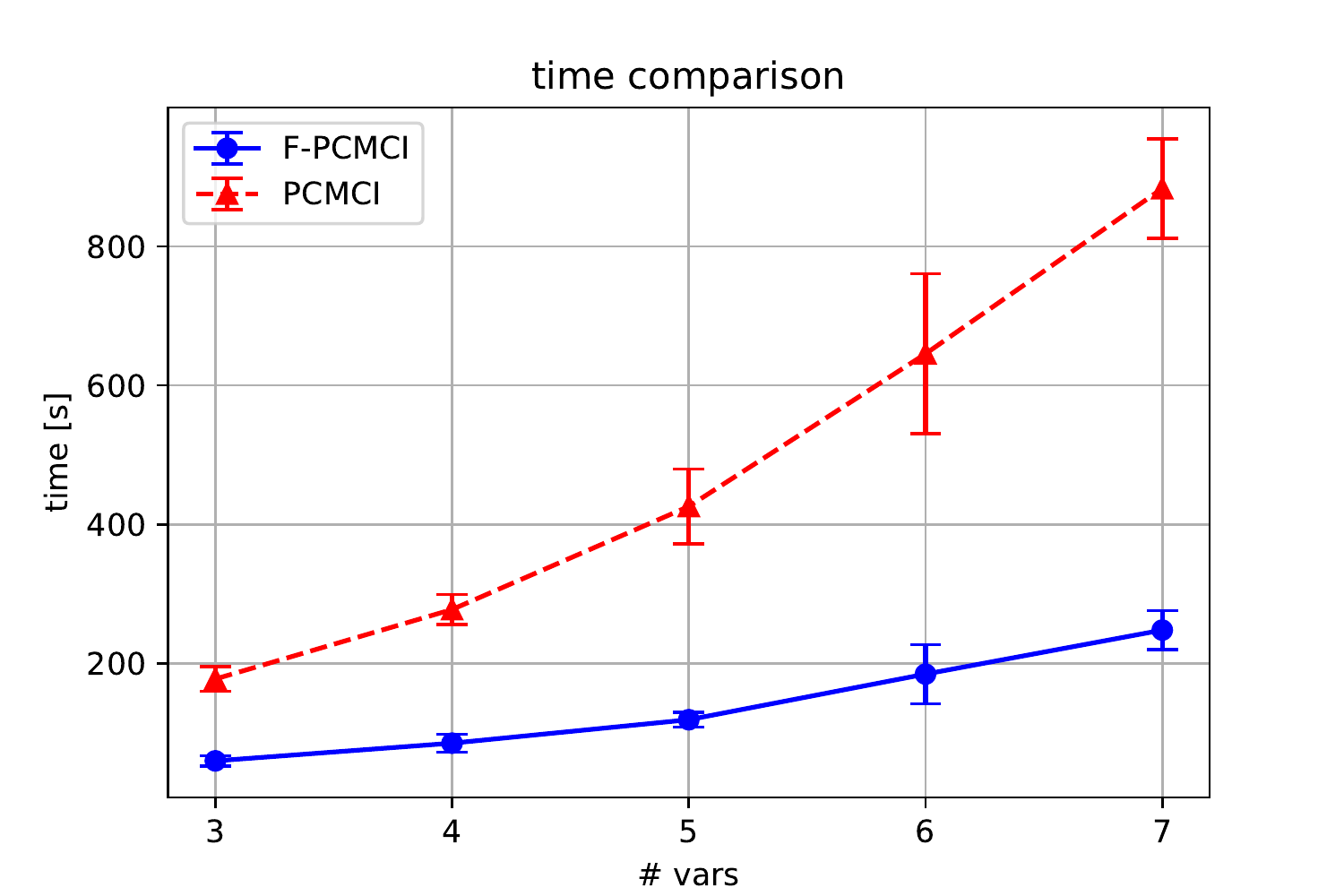}
    \label{fig:exp_S2_time}
\caption{
PCMCI~(red dashed) vs F-PCMCI~(blue) comparison for increasing number of variables based on SHD~(top line), F1-score~(central line) and execution time~(bottom line). The left and right column graphs are relative to $S_1$ and $S_2$, respectively. Points and error bars represent mean and standard deviation over 10 run tests.}
\label{fig:exp_toy}
\end{figure}

The evaluation of our F-PCMCI approach was carried out using the two systems of equations $S_1$ and $S_2$ defined in Sec.~\ref{subsec:toyproblem_appr}, with a number of variables ranging between 3 to 7, using time-series with 1500 data samples. For each system configuration, we performed 10 run tests with different random coefficients, using as evaluation metrics the mean Structural Hamming Distance~(SHD), the mean F1-score and the mean execution time (in seconds) over all the tests. 
Fig.~\ref{fig:exp_toy} shows a comparison between the results of the causal discovery performed using PCMCI~(red dashed lines) and our F-PCMCI~(blue lines). It clearly shows that F-PCMCI outperforms PCMCI in terms of accuracy and execution time, since the latter suffers when the number of variables increases. Our approach, instead, maintains a reasonably low execution time even for $S_2$, which has a high number of potential links to be checked due to the dependency time-lag of 2. 

Note that the time complexity of the PCMCI algorithm depends on the number of variables involved in the analysis and on the type of relationships those variables have with each other. This can be seen in Fig.~\ref{fig:exp_toy}~(bottom-left) relatively to $S_1$, where, for example, the time increase when the 6th variable $X_5$ was included is almost ``linear'', since it is just noise not connected to any other variable. Because of this, its addition makes the PC part of PCMCI to check more links, but the MCI part will not perform any further checks on that variable. Moreover, the time complexity depends also on the coefficients $c$, randomly chosen for each configuration. If they are very small, the algorithm is not able to detect the link (since the noise hides it), and therefore the time complexity does not increase exponentially as expected. On the other hand, when the coefficients are large, we would notice a clearer and bigger time difference between consecutive cases.


\subsection{Evaluation on fMRI Synthetic Data}
\begin{figure*}[t]
\centering 
    \includegraphics[trim={4.5cm 2cm 4.2cm 2cm}, clip, width=0.32\textwidth]{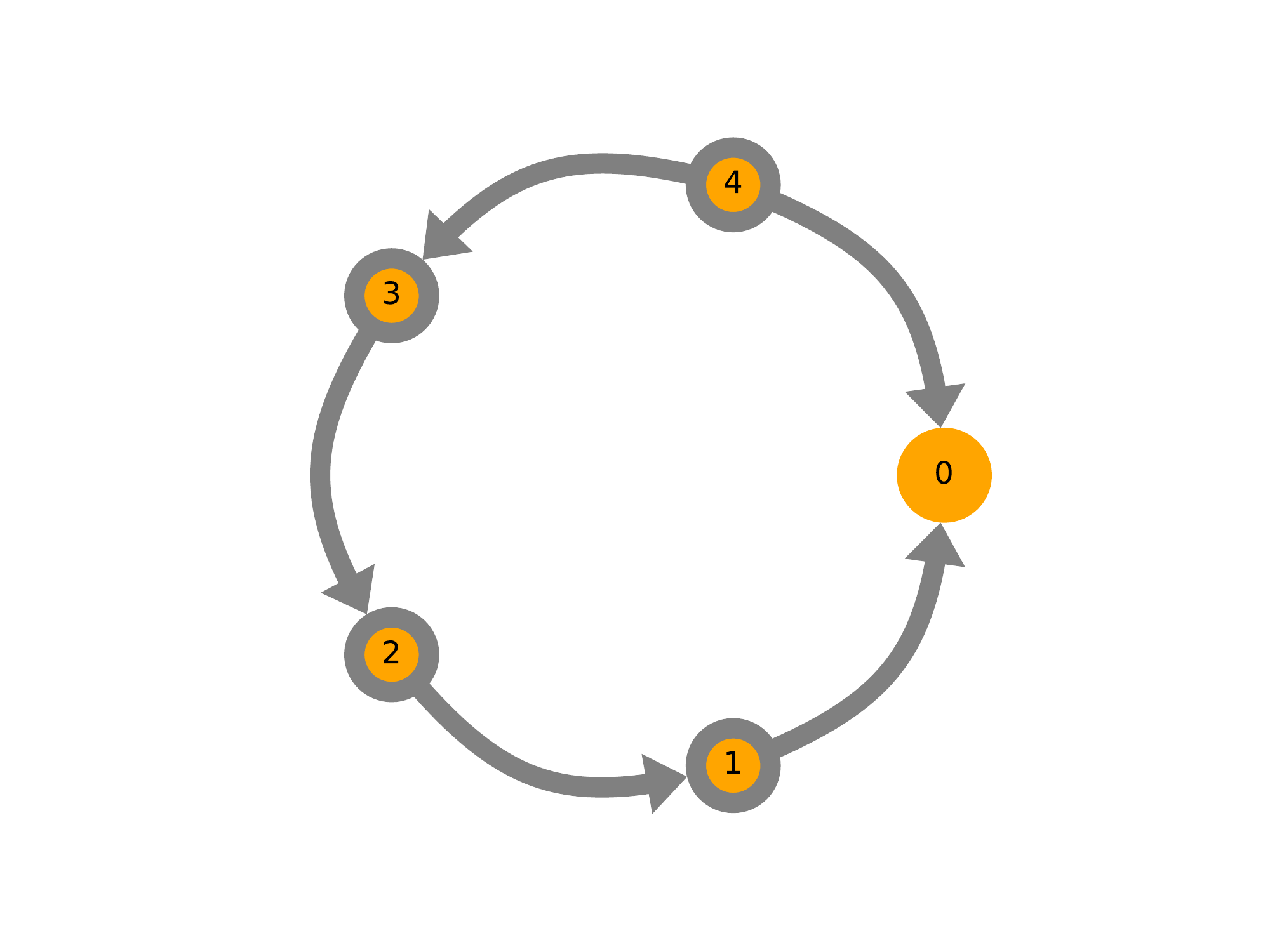}
    \label{fig:exp_FMRI_GT}
    \includegraphics[trim={4.5cm 2cm 4.2cm 2cm}, clip, width=0.32\textwidth]{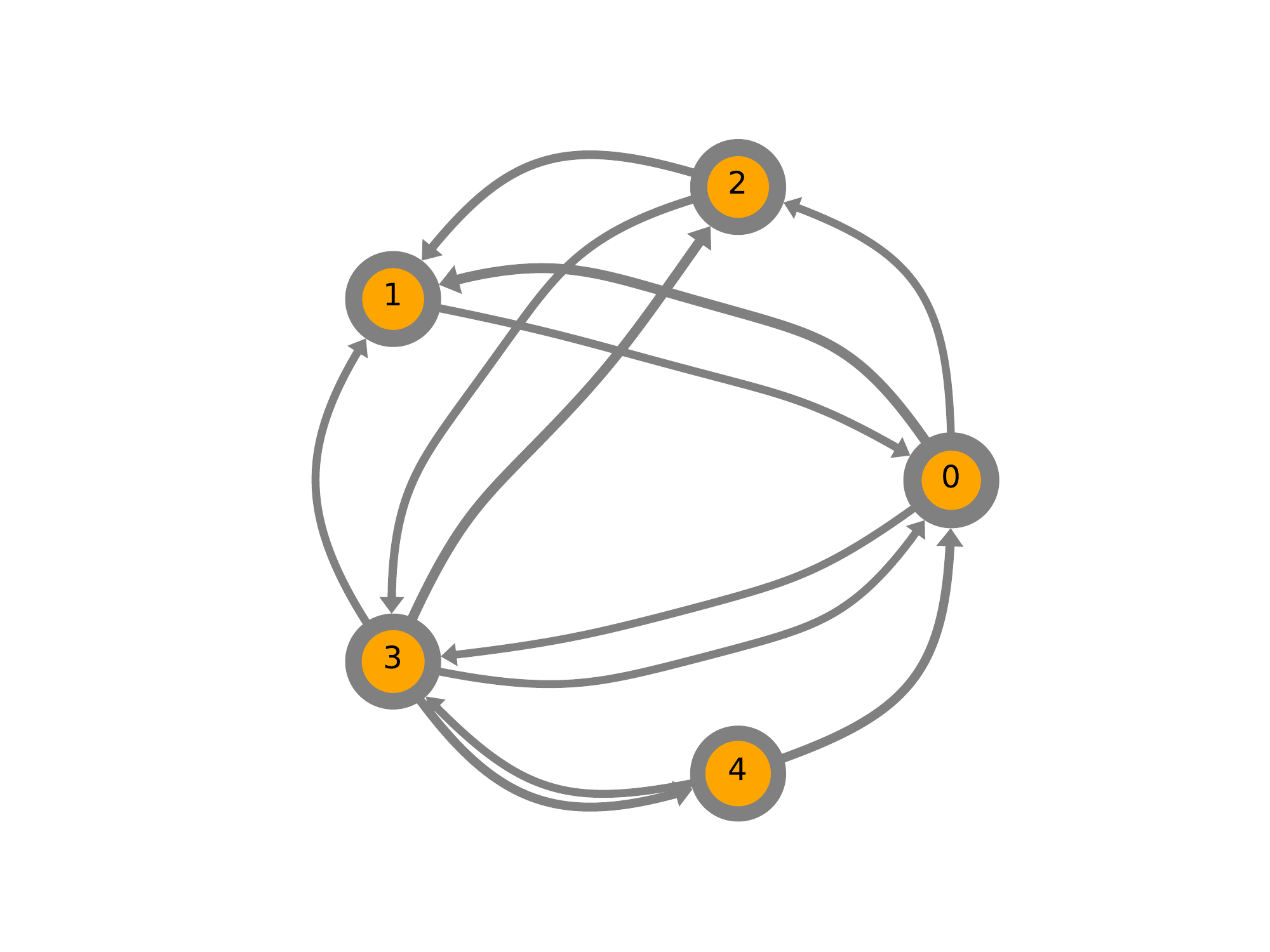}
    \label{fig:exp_PCMCI_FMRI_dag}
    \includegraphics[trim={4.5cm 2cm 4.2cm 2cm}, clip, width=0.32\textwidth]{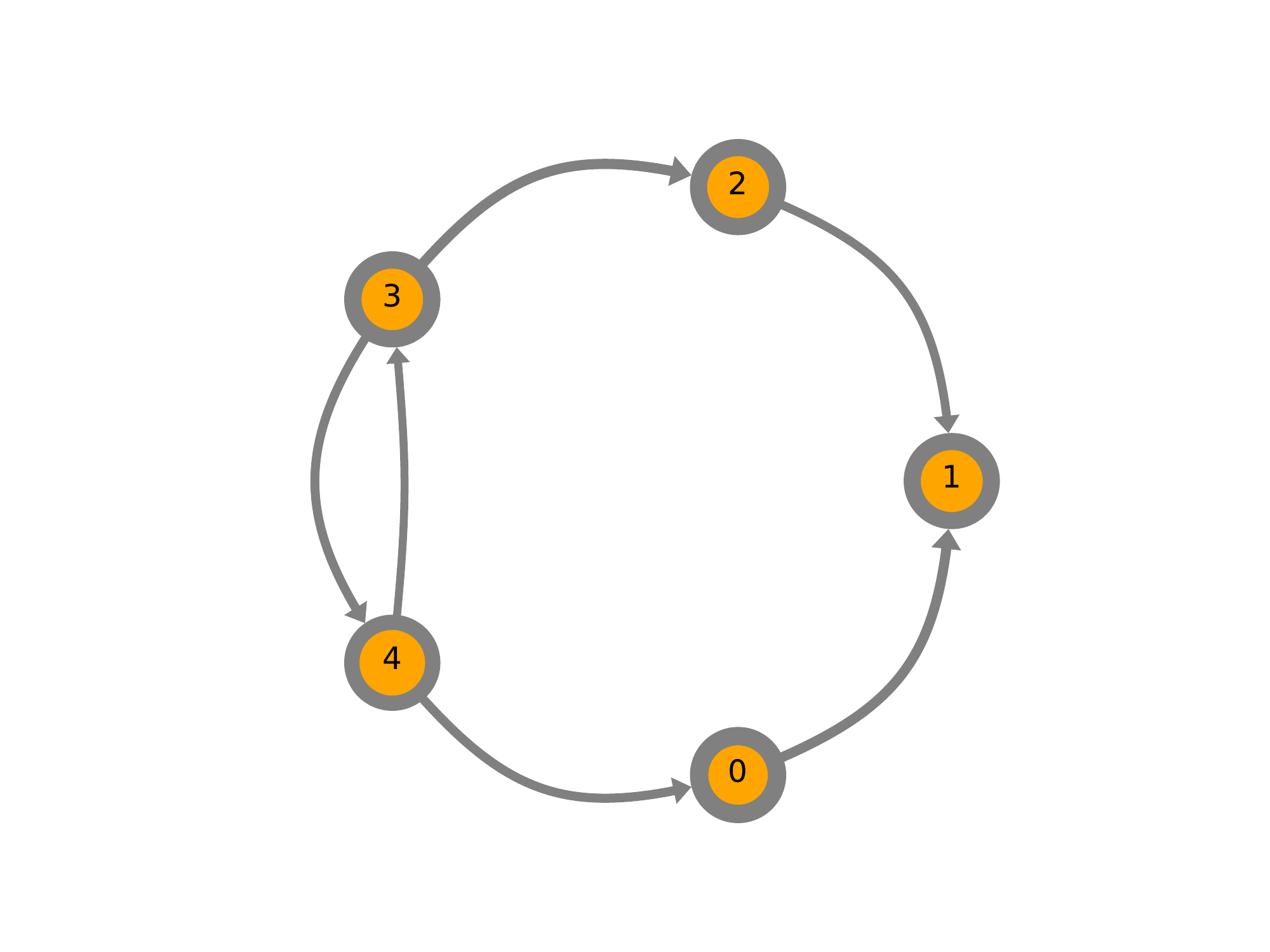}
    \label{fig:exp_FPCMCI_FMRI_dag}
\caption{
PCMCI vs F-PCMCI comparison using FMRI time-series data. (left) Ground-truth causal model; (centre) Causal model derived by the PCMCI; (right) Causal model derived by the F-PCMCI.}
\label{fig:exp_FMRI_dag}
\end{figure*}

\begin{wraptable}{r}{0.45\textwidth}
\vspace{-12pt}
\centering
\begin{tabular}{l|c|c|c}
       & SHD        & F1-Score      & Time                 \\ \hline
PCMCI  & 8          & 0.69          & 90'50"      \\
F-PCMCI & \textbf{4} & \textbf{0.80} & \textbf{38'52"}
\end{tabular}
\caption{
PCMCI vs F-PCMCI comparison on FMRI data, based on SHD, F1-Score and execution time.}
\label{table:exp_FMRI_dag}
\vspace{-20pt}
\end{wraptable}
We run a PCMCI vs F-PCMCI comparison on a fMRI dataset consisting of 5 time-series variables, which represent functional ``nodes'' (e.g., spatial ROIs or ICA maps), each one with 5000 samples. The causal analysis was based on a maximum time lag dependency equal to 1. As in the toy problem
evaluation of Sec.~\ref{subsec:toy_results}, the evaluation was based on SHD, F1-score, and execution time. Fig.~\ref{fig:exp_FMRI_dag} shows the comparison between the three causal models: (left)~ground-truth, (centre)~causal model generated by PCMCI, and (right)~causal model generated by F-PCMCI. From the figure it is already clear that PCMCI generates various spurious links between variables, instead the F-PCMCI's output is closer to the ground-truth causal model. These considerations are confirmed by the results in Table~\ref{table:exp_FMRI_dag}, which shows the benefits of our approach in terms of accuracy (lower SHD and higher F1-score) and execution time (two times faster than PCMCI).

\subsection{Real-world Experiment}
\begin{wrapfigure}{r}{0.4\textwidth}
  \vspace{-13pt}
  \centering
  \includegraphics[scale=0.19]{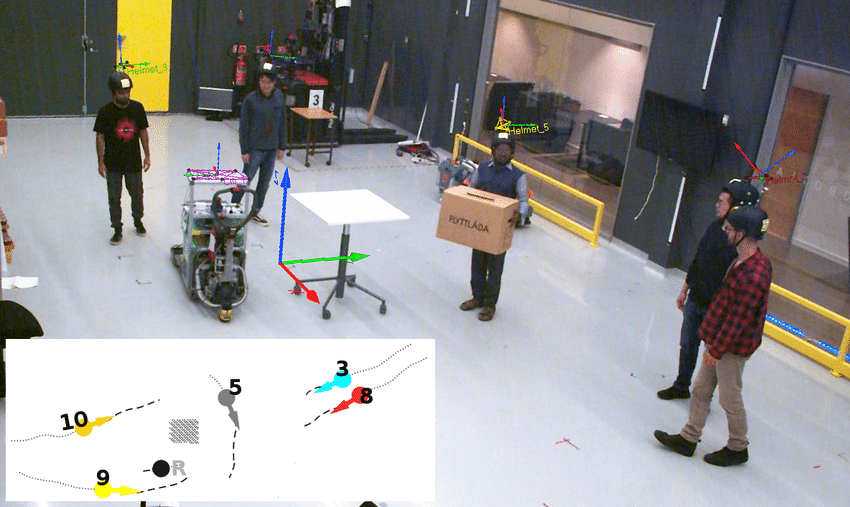} \label{fig:thor}
  \includegraphics[trim={6.4cm 1.6cm 6.3cm 1.8cm}, clip, scale=0.455]{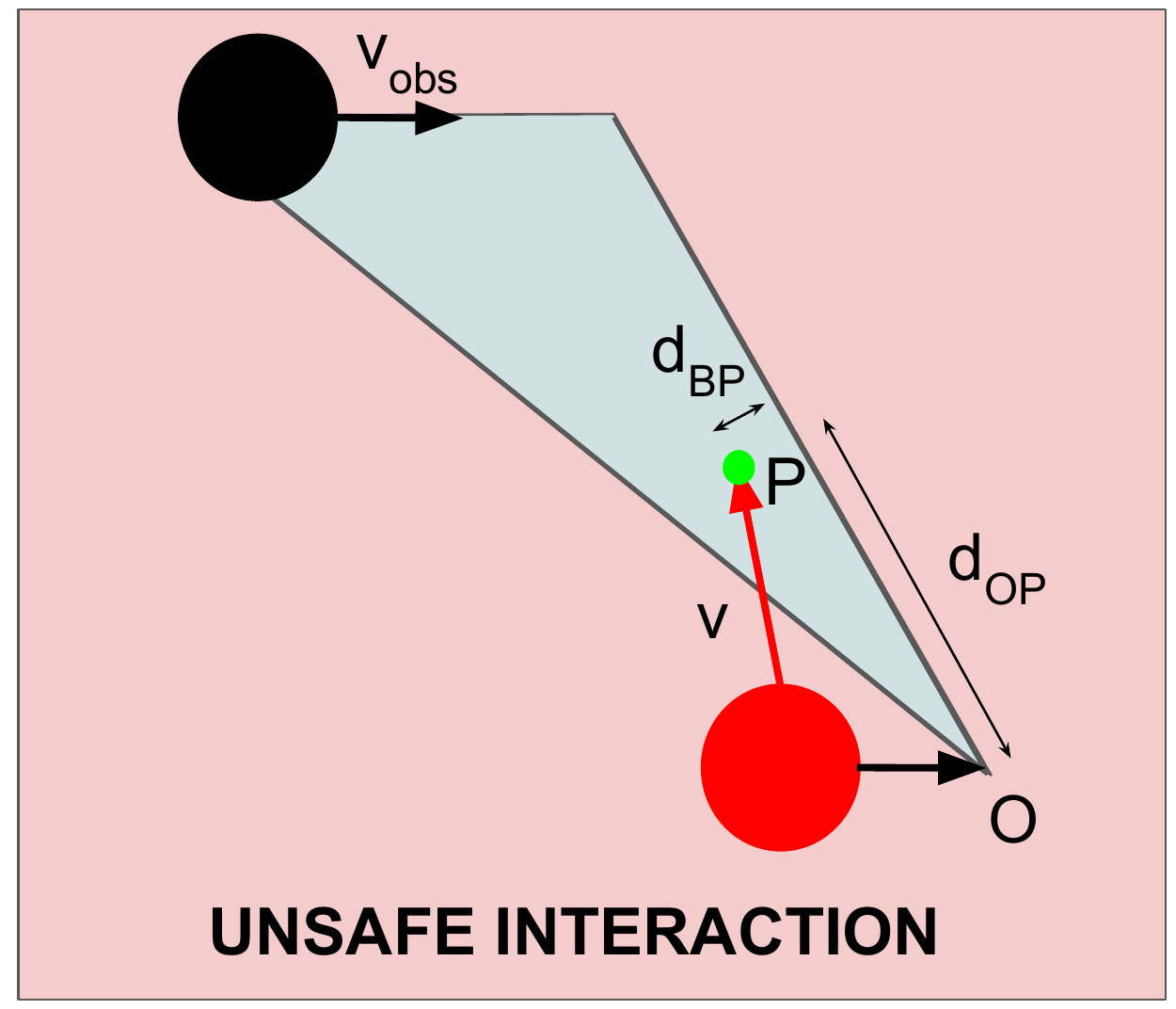}
  \label{fig:VO}
  \caption{(Top) Image from TH{\"O}R dataset. (Bottom) Risk analysis performed between the selected agent (red) and obstacle (black).}
  \label{fig:exp_setup}
  \vspace{-20pt}
\end{wrapfigure}

\textbf{Data Processing:}
From the TH{\"O}R dataset, we extracted the $x$-$y$ positions of each agent and derived from them all the quantities explained in the following paragraph. We used this dataset to analyse human spatial interactions, as described in Sec.~\ref{subsec:realworld_appr}, since it provides a wide variety of interactions between humans, robot, and static objects~(Fig.~\ref{fig:exp_setup} top). 
However, the original sampling frequency ($100~Hz$) of the dataset is very high for the type of scenario (people walking and carrying packages in a warehouse-like environment), leading to very long time-series with small differences between consecutive samples. To perform causal discovery on the original data, we would have to consider a number of possible time lag dependencies~$\gg 1$, making the analysis very slow and inefficient. The dataset was therefore subsampled using an entropy-based adaptive-sampling~\citep{Aldana-Bobadilla2015} and an additional variable size windowing approach to reduce the number of samples. This strategy allowed us to keep a maximum time-lag of 1.

In order to represent human spatial interactions, for each agent we considered 8 variables suitable for the application, which were then used in the causal analysis. These are the following:
\begin{itemize}
  \setlength{\itemsep}{1pt}
  \setlength{\parskip}{0pt}
  \setlength{\parsep}{0pt}
    \item[1)] $d_g$ -- distance between the current position of the agent and its goal;
    \item[2)] $v$ -- velocity of the selected agent;
    \item[3)] $risk$ -- risk of collision with other agents (explained below);
    \item[4)] $\theta$ -- orientation of the selected agent;
    \item[5)] $\theta_g$ -- angle between the current position and goal of the selected agent;
    \item[6)] $\omega$ -- angular velocity of the selected agent;
    \item[7)] $g_{seq}$ -- sequence of goal positions reached by the selected agent;
    \item[8)] $d_{obs}$ -- distance between the current position of the selected agent and the closest obstacle.
\end{itemize}

\noindent\textbf{Risk Evaluation:}
As proposed in~\citep{castri2022causal}, the risk of collision for a selected agent is evaluated using the Velocity Obstacle~(VO) strategy with respect to the closest obstacle~(Fig.~\ref{fig:exp_setup}~bottom). The risk is a function of two parameters: $d_{OP}$, which measures the time available for the selected agent to avoid a collision, and $d_{BP}$, which indicates the steering effort by the same agent to avoid such collision. This risk is significant only if the closest obstacle is within a certain threshold distance ($d_{thres}$=$1.5m$).
%
Its value is computed as follows:
    \begin{equation} \label{eq:risk}
    \begin{cases}
    risk = e^{d_{OP} + d_{BP} + v} \hspace{0.5cm} d_{obs} \leq d_{thres}\\
    risk = e^{v} \hspace{2.15cm} else
    \end{cases}
    \end{equation}
As shown in the equations, the $risk$ depends on the selected agent's velocity~$v$ and its proximity to the obstacle~$d_{obs}$.\\[0.1cm]
\noindent\textbf{Network Parameter Settings:}
As explained in Sec.~\ref{subsec:realworld_appr}, a LSTM-based encoder-decoder model was implemented for Multi-Output Multi-Step forecasting. We used a 70\%-10\%-20\% split of our time-series dataset for training, validation, and testing, respectively. 
To optimise the network, we perform a grid search to select the learning rate over \{0.01, 0.001, 0.0001\}, batch size over \{32, 64, 128\} and the number of LSTM cells per hidden layer over \{64, 128, 256, 512\} which lead to the best performance over the validation set. As result, for the training phase we adopted a learning rate of 0.0001, a batch size of 32 and a number of LSTM cells per hidden layer of 256 for both the encoder (2 hidden layers) and the decoder (2 hidden layers). Moreover, we set an observation window size of 32 (3.2sec) and forecasting window size of 48 (4.8sec) as in \citep{bartoli2018context,vemula2018social}.\\[0.1cm]
%
%
\noindent\textbf{Results}\\
This experiment was performed in order to evaluate our F-PCMCI on a real-world scenario of human spatial interaction. However, since in this case we do not have a ground-truth model, we assessed the correctness and accuracy of our causal model by looking at the prediction accuracy of the causality-augmented architecture explained in Sec.~\ref{subsec:realworld_appr}, which is also useful application in many real-world problems. First of all, we extracted the data from the TH{\"O}R dataset and derived the previously mentioned set of 8 variables to represent human spatial interactions. Then, the set is filtered and used to generate the causal model with our F-PCMCI.
After this initial step, we exploit the discovered causal model in the LSTM-based network to verify the improvement of its prediction accuracy with respect to a non-causal version of it. For comparison, we repeated the same process with the causal model generated by PCMCI.

\begin{figure}[t]
\centering 
    \includegraphics[trim={6.5cm 2.5cm 2.5cm 1.3cm}, clip, width=0.48\textwidth]{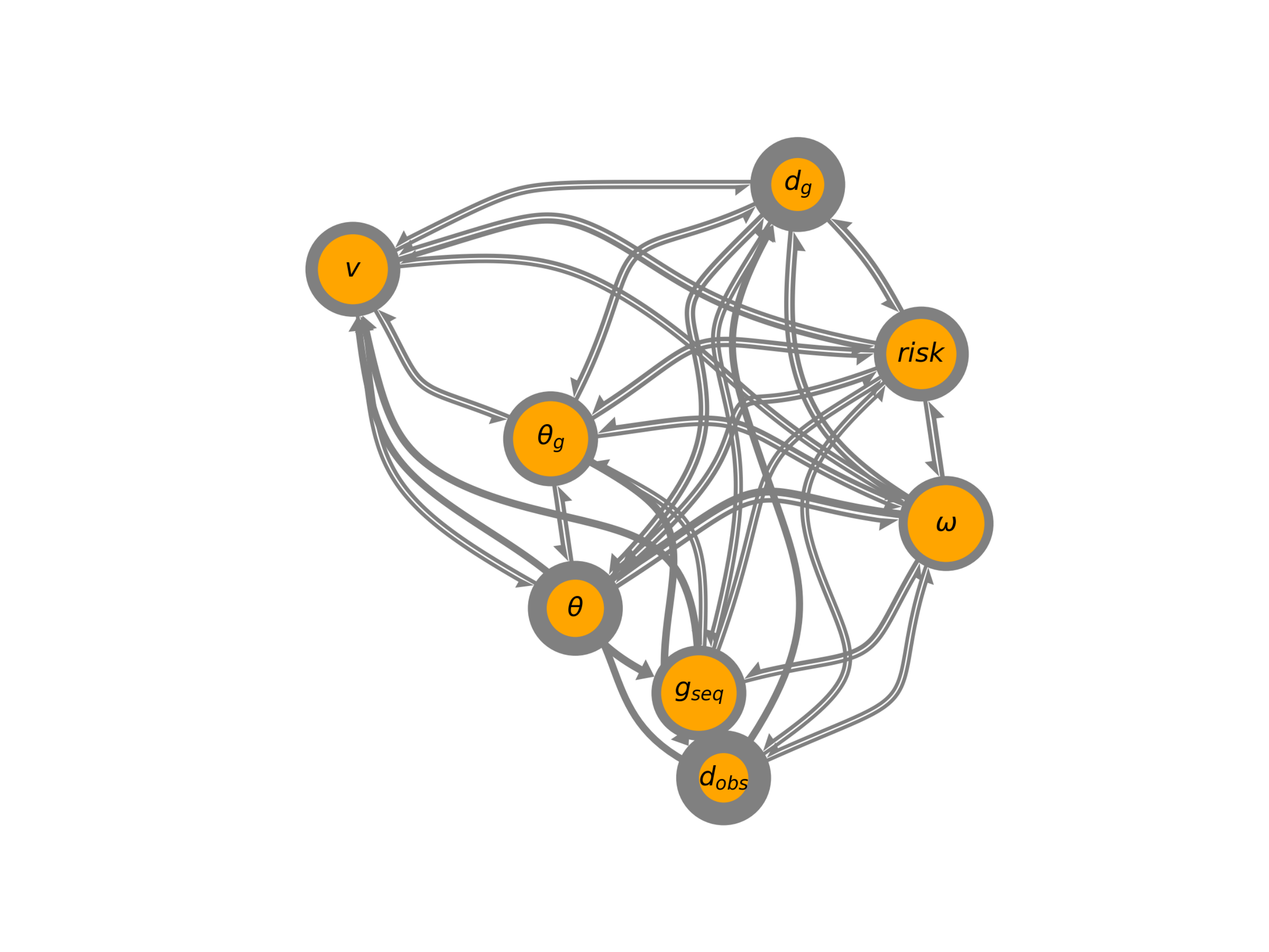}
    \label{fig:exp_PCMCI}
    \includegraphics[trim={5cm 2cm 2.5cm 1.3cm}, clip, width=0.48\textwidth]{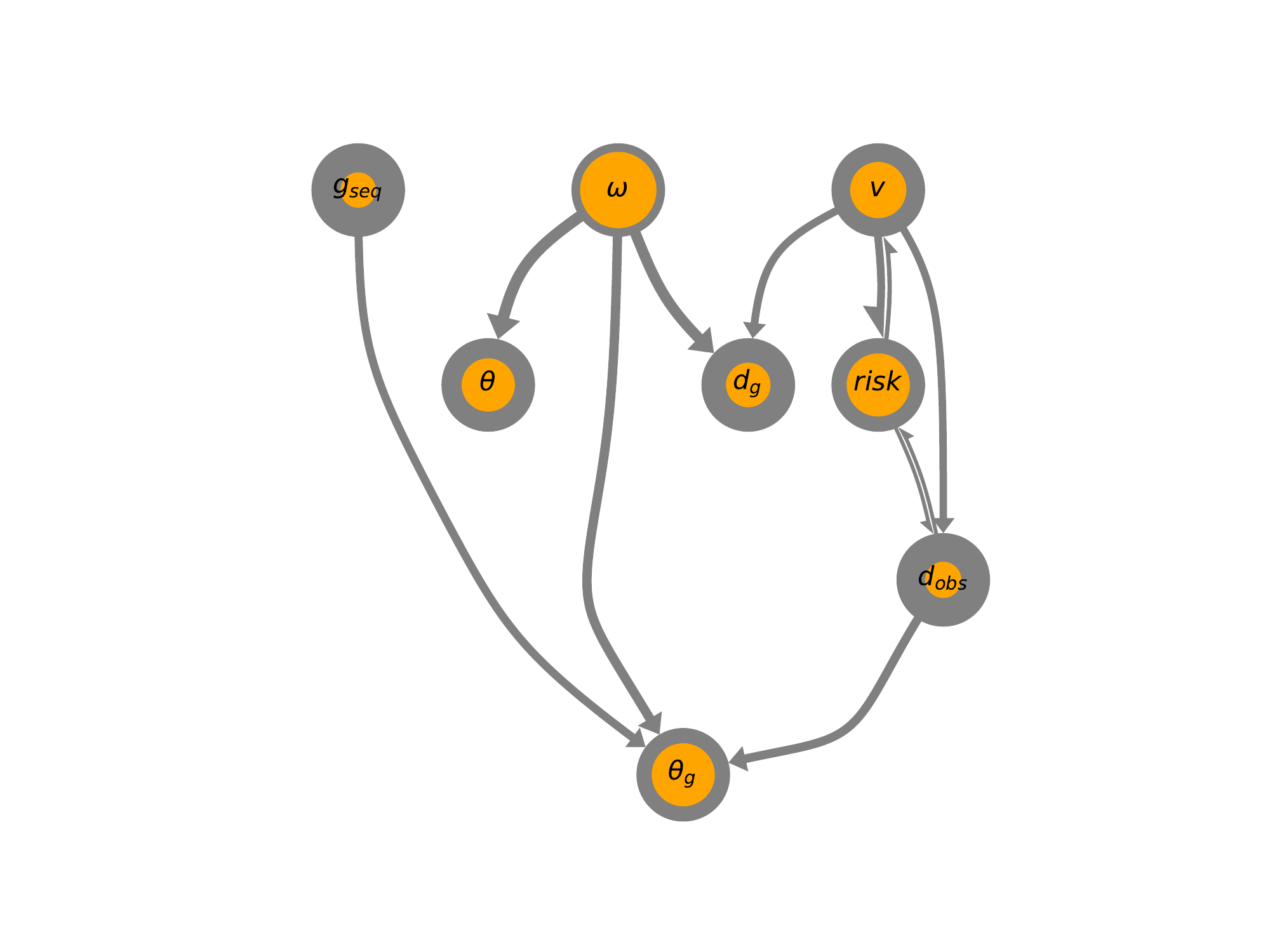}
    \label{fig:exp_FPCMCI}
\caption{
Causal models of the TH{\"O}R dataset using PCMCI~(left) and F-PCMCI (right). Arrows and borders of the nodes represent the strength of cross-causal and auto-causal dependencies, with stronger dependencies shown by thicker lines/borders. All dependencies have a 1-step lag time.}
\label{fig:exp_dag}
\end{figure}
Fig.~\ref{fig:exp_dag} shows the two causal models derived from PCMCI and F-PCMCI relatively to agent~11 of the TH{\"O}R dataset. As expected, due to the large number of variables and links, the PCMCI algorithm is affected by spurious links, which it is not able to filter out. On the other side, F-PCMCI provides a simpler and more realistic causal model, which includes the full set of variables (as PCMCI) but keeps only the most meaningful links between them, thanks to the TE-based filtering step. The execution time of the causal discovery confirmed our previous results: indeed PCMCI completed in 79'45", while the F-PCMCI's execution lasted only 17'33", i.e. more than 4~times faster (the machine used for the experiment is a Lenovo Legion~i7).

Lacking a ground truth model, we can only judge qualitatively its correctness. In particular, from Fig.~\ref{fig:exp_dag}, we can observe the following:\\
\begin{itemize}
  \setlength{\itemsep}{1pt}
  \setlength{\parskip}{0pt}
  \setlength{\parsep}{0pt}
    \item the $v \leftrightarrow risk \leftrightarrow d_{obs}$ links describe the risk of the observed agent due to the presence of other agents in the scenario. In particular, $risk$ depends on $v$ and is effective only when $d_{obs}$ is smaller than a certain threshold. On the other hand, the velocity $v$ depends on the $risk$ value. For example, if $risk$ is high, the observed agent could either stop or increase its velocity to avoid a possible collision. This explains also the causal relation between $d_{obs}$ and~$v$;
    \item the $\omega \rightarrow d_g \leftarrow v$ links indicates that the distance between the observed agent and its target position depends on its linear and angular velocities;
    \item the $\theta \leftarrow \omega \rightarrow \theta_g$ links refers to the orientation of the observed agent; also, when the target position sequence~($g_{seq}$) changes, the angle between the observed agent and the new goal~($\theta_g$) changes as well;
    \item the $d_{obs} \rightarrow \theta_g$ link explains the fact that a small distance between the observed agent and the obstacle leads to a change of the angle $\theta_g$.
\end{itemize}

We trained a new network for each agent in the scenario and test it on the other agents. To evaluate the quality of prediction we used the \textit{Normalised Mean Absolute Error}~(NMAE) and the \textit{Normalised Root Mean Square Error}~(NRMSE), defined as follows:
\begin{center}
\begin{minipage}[t]{.49\textwidth}
\begin{equation}
NMAE(y, \hat{y}) = \frac{\sum_{i=1}^{n}\frac{\lvert y_i - \hat{y_i}\rvert}{n}}{\sigma(y)}
\end{equation}
\end{minipage}
\begin{minipage}[t]{.49\textwidth}
\vspace{-5pt}
\begin{equation}
NRMSE(y, \hat{y}) = \frac{\sqrt{\sum_{i=1}^{n}\frac{(y_i - \hat{y_i})^2}{n}}}{\sigma(y)}
\end{equation}
\end{minipage}
\end{center}
where $y$ and $\hat{y}$ are the actual and the predicted values, respectively, and $\sigma(y)$ is the standard deviation of the actual value.
\begin{figure}[t]\centering
\includegraphics[trim={0.1cm, 0.1cm, 7.5cm, 13cm}, clip, width=\textwidth]{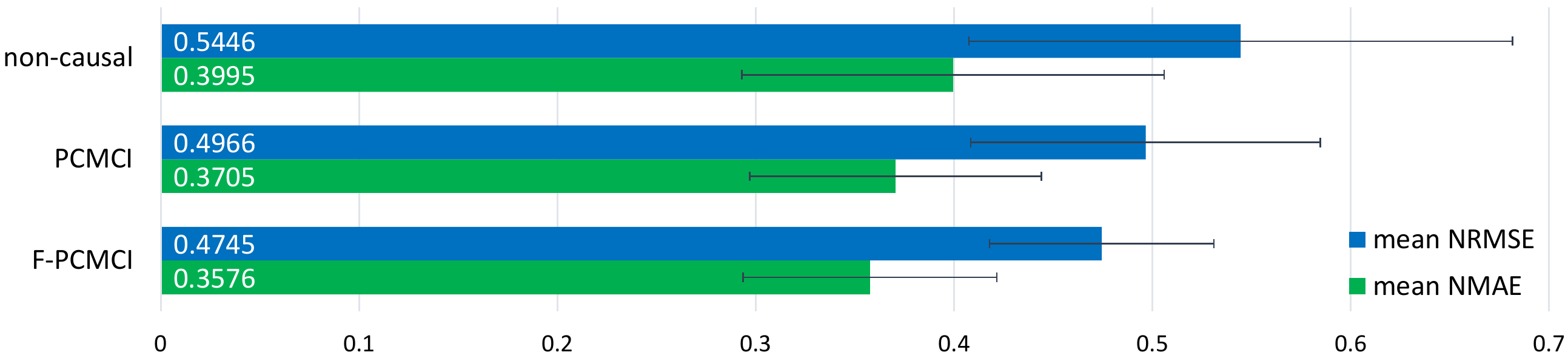}
\caption{Comparison between non-causal and causal prediction (with both PCMCI and F-PCMCI) using mean NMAE and mean NRMSE across all the agents in the scenario. White numbers and error bars indicate mean and standard deviation, respectively.}
\label{fig:exp_bar_chart}
\end{figure}
Fig.~\ref{fig:exp_bar_chart} reports the comparison between prediction accuracy in the three cases: non-causal prediction, PCMCI-based prediction, and F-PCMCI-based prediction. The NMAE and NRMSE values are computed for each selected agent and then averaged. The figure clearly shows that the knowledge of the causal model helps to obtain a more accurate prediction. Moreover, since both errors are lower for the F-PCMCI's case compared to the PCMCI's one, we can conclude that our approach produces a better and more useful causal model.

\section{Conclusion} \label{sec:conclusion}
In this work, we extended and improved a state-of-the-art causal discovery algorithm, PCMCI, embedding an additional feature-selection module based on transfer entropy. The proposed method was initially evaluated on two toy problems and on synthetic data of brain networks, for which the ground-truth models were known a priori, to verify the correctness of the approach. It was then tested on a real-world robotics dataset with large-scale time-series of human trajectories. We showed that our approach significantly improves the PCMCI causal discovery method in terms of accuracy and computational efficiency. This leads to better and faster causal discovery of dynamic models from real-world sensor data.

Future work will be devoted to augment our FPCMCI with a strategy, inspired by \citep{yao2022temporally,lippe2022citris}, to start the causal discovery process from a set of variables which are not known a priori, but automatically extracted from the scenario. Moreover, we plan to use the robot as an active agent that performs interventions to discover new causal links, as required by~\citep{lachapelle2022disentanglement,wang2022causal,lippe2022citris}, for example exploiting its potential influence on nearby people, with a special interest on industrial and intralogistics applications.
%

\acks{This work has received funding from the European Union's Horizon 2020 research and innovation programme under grant agreement No 101017274 (DARKO).}

\bibliography{sections/references}






\end{document}